\documentclass{article}
\usepackage{spconf,amsmath,graphicx}
\usepackage[utf8]{inputenc} 
\usepackage[T1]{fontenc}    
\usepackage{hyperref}       
\usepackage{url}            
\usepackage{booktabs}       
\usepackage{amsfonts}       
\usepackage{nicefrac}       
\usepackage{microtype}      
\usepackage{graphicx}
\usepackage{threeparttable}
\usepackage[vlined,boxed,commentsnumbered,ruled,linesnumbered]{algorithm2e}
\usepackage{amsmath,amssymb}
\usepackage{multirow}
\usepackage{subfigure}
\usepackage{comment}
\usepackage{tablefootnote}
\usepackage{stfloats}
\usepackage[table]{xcolor}
\usepackage{amsmath}
\usepackage[linesnumbered, ruled]{algorithm2e}
\usepackage{bm}

\usepackage{colortbl}

\DeclareUrlCommand\url{\color{blue}}

\title{GMTR: Graph Matching Transformers}

\name{Jinpei Guo, Shaofeng Zhang, Runzhong Wang, Chang Liu, Junchi Yan$^*$\thanks{$^*$Corresponding author. Work was supported by National Key Research and Development Project (2020AAA0107600) and NSFC (62222607).}}
\address{MOE Key Lab of Artificial Intelligence, Shanghai Jiao Tong University, Shanghai, China\\	
}

\begin{document}
%
\maketitle
%
\begin{abstract}

Vision transformers (ViTs) have recently been used for visual matching. The original grid dividing strategy of ViTs neglects the spatial information of the keypoints, limiting the sensitivity to local information. We propose \textbf{QueryTrans} (Query Transformer), which adopts a cross-attention module and keypoints-based center crop strategy for better spatial information extraction. We further integrate the graph  attention module and devise a transformer-based graph matching approach \textbf{GMTR} (Graph Matching TRansformers) whereby the combinatorial nature of GM is addressed by a graph transformer GM solver. On standard GM benchmarks, GMTR shows competitive performance against the SOTA frameworks. Specifically, on Pascal VOC, GMTR achieves $\mathbf{83.6\%}$ accuracy, $\mathbf{0.9\%}$ higher than the SOTA. On SPair-71k, GMTR shows great potential and outperforms most of the previous works. Meanwhile, on Pascal VOC, QueryTrans improves the accuracy of NGMv2 from $80.1\%$ to $\mathbf{83.3\%}$, and BBGM from $79.0\%$ to $\mathbf{84.5\%}$. On SPair-71k, it improves NGMv2 from $80.6\%$ to $\mathbf{82.5\%}$, and BBGM from $82.1\%$ to $\mathbf{83.9\%}$. Code is available at: https://github.com/jp-guo/gm-transformer.

\end{abstract}
 

\section{Introduction and Related Work}\vspace{-7pt}
\label{sec:intro}

Since the debut of Vision Transformer (ViT)~\cite{vit}, variant transformers~\cite{liu2021swin, setr} have been devised to handle different vision tasks (e.g., DETR~\cite{detr} on detection and SETR~\cite{setr} on segmentation) and outperform  ConvNets. However, when applying transformers to visual graph matching, a standard spatial-sensitive task involving keypoints matching, the grid dividing design in ViTs omits the keypoints spatial information and therefore poses a significant limitation to the performance of transformers. Hence, exploring the potential of transformers in visual graph matching remains open.

Visual GM extracts keypoint features and model their relation from images, and aims to find the correspondence between two or multiple images, which is essentially NP-hard with wide applications in vision. Most existing visual GM methods~\cite{ZanfirCVPR18, WangICCV19, WangPAMI22, YuICLR20} adopt the ConvNets (e.g., VGGs and ResNets) to extract keypoint information, where CNNs are shown to be less competitive to \textbf{relational modeling}~\cite{graphtransformer-2020} (e.g. convolution for sequential data). To solve the two bottlenecks (less spatial sensitiveness and relational modeling), we propose QueryTrans (Query Transformer), integrating transformers in visual GM and solving the spatial-insensitive problem for ViTs. We split the image into patches (raw patches). For each keypoint, we crop a region around the keypoint (key patches) with the same size of raw patches followed by the vision transformer. 

\begin{figure}[tb!]
    \begin{center}
        \includegraphics[width=\columnwidth]{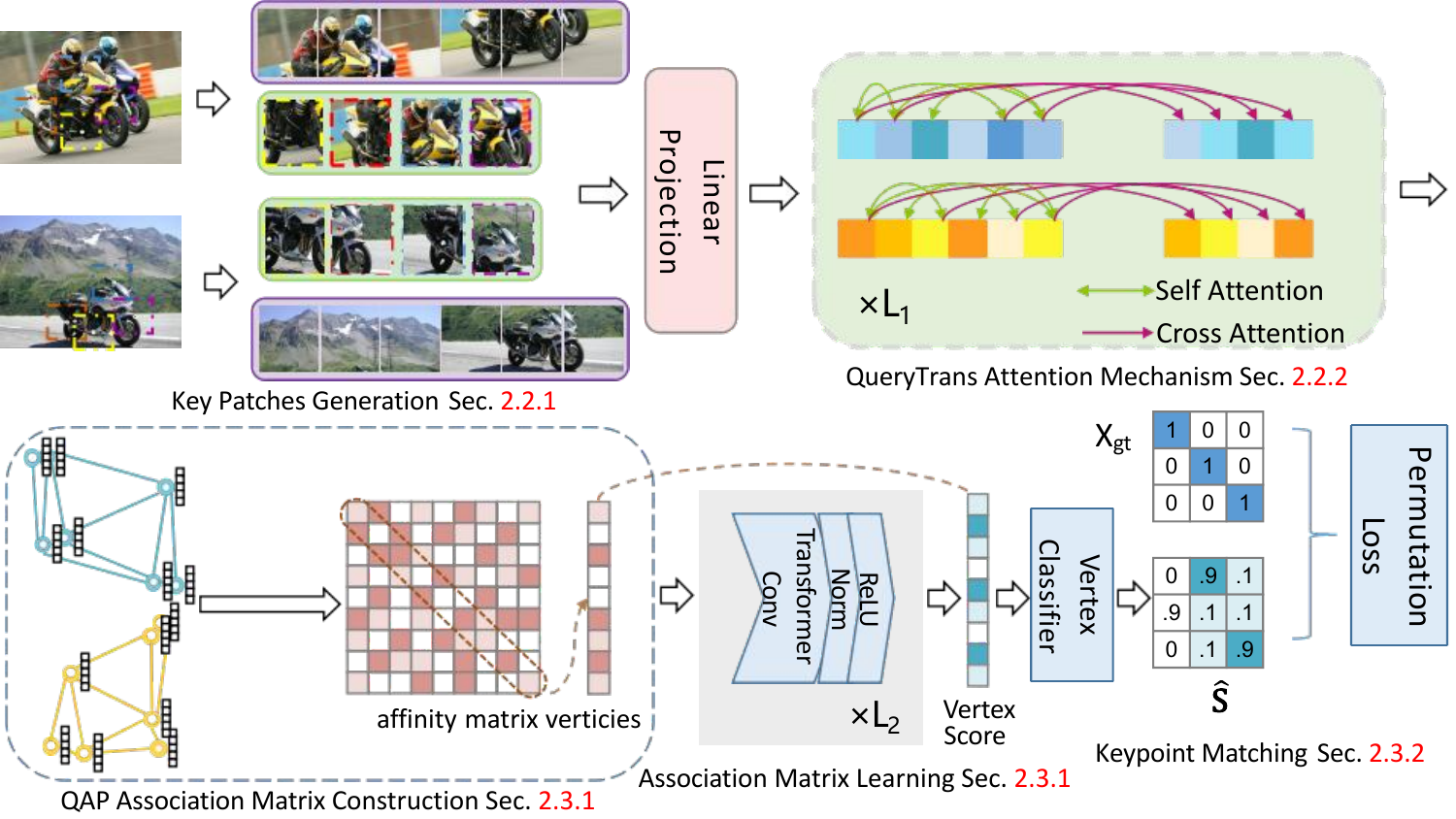}
    \end{center}
    \vspace{-20pt}
    \caption{Overview of GMTR. The frontend QueryTrans extracts image features via generated raw patches and key patches (region crops around keypoints), followed with self-attention and cross-attention layers. The affinity matrix as input instance, is then tackled by the backend transformer. All modules are end-to-end trained with the permutation loss~\cite{WangICCV19}.
    }
    \label{fig:framework}
    \vspace{-15pt}
\end{figure}

Previous works have applied the transformers to image matching and show promising potential. While graph matching refers to keypoint to keypoint matching of graph structure, \cite{superglue, LoFTR_CVPR21} focus on dense pixel to pixel matching of images, which show few strengths on graph structure. Instead of straightforwardly borrowing transformers in our approach, we exploit the graph structure and design a cross attention mechanism between raw patches and key patches to decrease the bias for better localized keypoints information. Following \cite{WangPAMI22} that treats the NP-hard quadratic assignment problem (QAP) solving as a vertex classification task, we naturally replace GNNs with transformers to exploit the mutual information in the graph structure and explore their potential in solving combinatorial optimization problems. By introducing the key patches, cross attention mechanism, and the transformer GM solver, GMTR achieves new SOTA results on standard visual graph matching datasets Pascal VOC (with Berkeley annotations~\cite{bourdev2009poselets}) and exhibits great potential on  SPair-71k~\cite{min2019spair}. 

\textbf{The contributions are summarized as follows:} 1) We propose a transformer-based method namely Graph Matching TRansformers (GMTR) to improve the relational modeling and further expoit the property of graph structure. 2) We devise techniques including raw/key patch cross attention and center region crop for better keypoint feature extraction, improving the sensitivity over original ViTs. 3) We conduct comprehensive experiments on popular GM benchmarks which prove the strengths of our proposed methods on all the datasets.

\begin{figure}[tb!]
    \begin{center}
        \includegraphics[width=0.9\columnwidth]{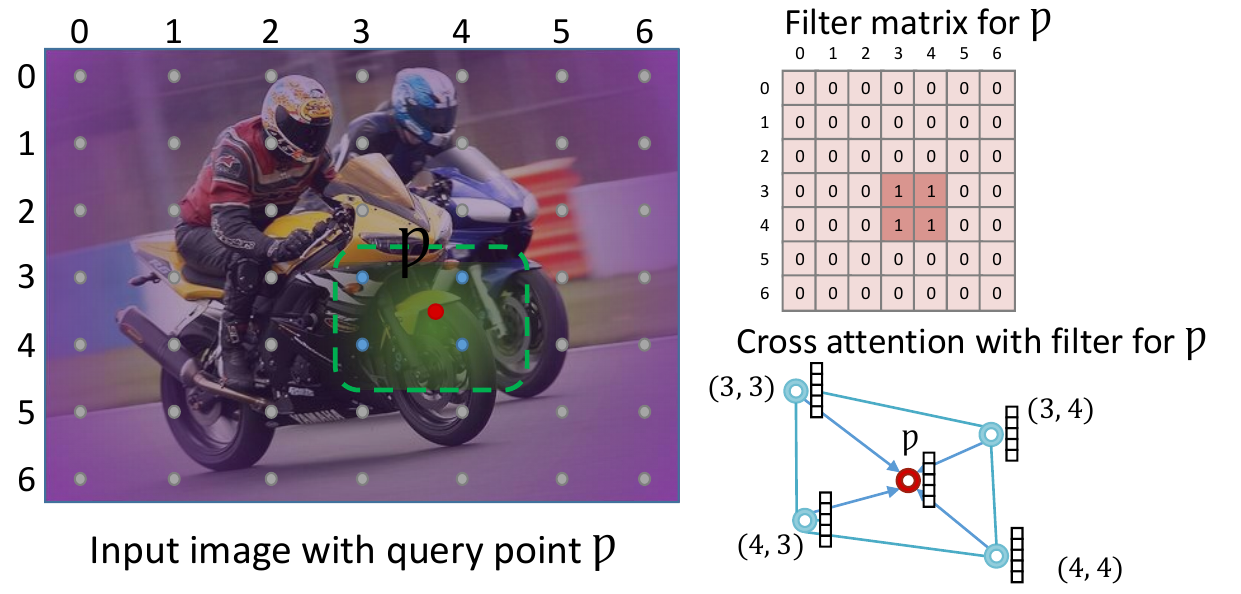}
    \end{center}
    \vspace{-20pt}
    \caption{Illustration of filter matrix $\mathbf{D}$ used in our devised QueryTrans for the query point $\mathcal{P}$ for keypoint feature learning, which conducts message passing with neighboring $4$ patches. Each grey point the center of raw patches, red point denotes the center of query patches, blue points the center of neighboring raw patches for query patches. See details in Sec.~\ref{sec:att_filter}.}
    \vspace{-10pt}
    \label{fig:illufilter}
\end{figure}

\vspace{-7pt}
\section{The Proposed Method}
\vspace{-7pt}


\subsection{Preliminaries: Vision Transformers}
\vspace{-7pt}

For image $\mathbf{x} \in \mathbb{R}^{C \times H \times W}$, where $H \times W$  is the resolution of the image and $C$ is the number of channels, ViT~\cite{vit} treats the image $\mathbf{x}$ as a sequence composed of non-overlapping patches $\{\mathbf{x}^{(i)} \in \mathbb{R}^{CP^2}\}_{i=1}^{N}$, where each patch has a fixed $P\times P$ resolution. Then, the patches are linearly transformed to $D$-dimensional patch embeddings $\mathbf{z}^{(i)} = \mathbf{Ex}^{(i)} + \mathbf{W}_{pos}^{i} \in \mathbb{R}^{D}$, where $\mathbf{E} \in \mathbb{R}^{D \times CP^2}$ is the linear projection and $\mathbf{W}_{pos} \in \mathbb{R}^{D}$ is the positional embedding for the $i$-th patch. A $[CLS]$ token $\mathbf{z}^{[CLS]} \in \mathbb{R}^{D}$ is subsequently preposed to the patch sequence to extract global information, so the resulting input sequence is represented as 
$\mathbf{z} = [\mathbf{z}^{[CLS]}, \mathbf{z}^{(1)}, \mathbf{z}^{(2)}, \cdots, \mathbf{z}^{(N)}]$. Then, ViT uses a transformer encoder to generate both image-level ($[CLS]$ token) and patch-level (other tokens). In line with~\cite{vit}, we use $f_{\theta}$ to denote the process of a ViT parameterized by $\theta$:
\begin{equation}
\begin{aligned}
\label{eqn:extract}
    f_{\theta}(\mathbf{x}) &= f_{\theta}\left(\left [\mathbf{z}^{[CLS]}, \mathbf{z}^{(1)}, \mathbf{z}^{(2)}, \cdots, \mathbf{z}^{(N)}\right ]\right) \\
    & = \left [ f_{\theta}^{[CLS]}(\mathbf{x}), f_{\theta}^{(1)}(\mathbf{x}), f_{\theta}^{(2)}(\mathbf{x}), \cdots, f_{\theta}^{(N)}(\mathbf{x})\right ],
\end{aligned}
\end{equation}
where $f_{\theta}^{[CLS]}(x)$ and $f_{\theta}^{(i)}(\mathbf{x})$ are the representations of the whole image and the $i$-th patch, respectively.


\vspace{-7pt}
\subsection{Frontend: Query Transformer}\vspace{-7pt}
\subsubsection{Key patches generation}
\label{sec:patches}
\vspace{-7pt}
ViT's spatial insensitivity can limit its ability to extract keypoint information, which may negatively impact its performance in GM tasks. As a remedy, we propose creating patches around keypoints to improve expressiveness. Specifically, as shown in top left of Fig.~\ref{fig:framework}, for each keypoint, we crop a region with the same size of raw patches (e.g., $8\times 8, 16\times 16$) followed by the patch embedding layer, as formulated by:
\begin{equation}
\label{eqn:patchembed}
    \left [ \mathbf{p}^{[(N+1), \cdots, (N+K)]} \right ] = \operatorname{PatchEmbed} \left (\mathbf{x}^{[(N+1), \cdots, (N+K)]} \right )
\end{equation}
where $\mathbf{p}^{[(N+1), \cdots, (N+K)]}$ means a sequence composed of the key patches embeddings. For distinguish, we call the patches generated by original ViT as \textbf{raw} patches and those patches generated by region crop around keypoints as \textbf{key} patches. Then, the input embeddings can be formulated as:
\begin{equation}
\label{eqn:input}
    \mathbf{z} = \left [ \mathbf{z}^{[CLS]}, \underbrace{\mathbf{z}^{(1)}, \mathbf{z}^{(2)}, \cdots, \mathbf{z}^{(N)}}_{\text{raw patches}}, \underbrace{\mathbf{z}^{(q_1)}, \mathbf{z}^{(q_2)},\cdots, \mathbf{z}^{(q_Q)}}_{\text{key patches}} \right ].
\end{equation}
Note that $\mathbf{z}^{[CLS]}$ and raw patches $[\mathbf{z}^{(1)}, \mathbf{z}^{(2)}, \cdots, \mathbf{z}^{(N)}]$ will be equipped with positional embeddings.


\subsubsection{Cross attention mechanism with filter module}
\vspace{-7pt}
\label{sec:att_filter}
Instead of directly feeding the whole sequence (namely $[CLS]$ token, raw patches, query patches) to the transformer encoder, which may hurt the performance of the finetuning on GM task (as only $[CLS]$ token and raw patches are fed to the transformer for pretraining), we modify the attention mechanism as follows. Specifically, given one $\mathbf{z}^{[CLS]}$ token, raw patches sequence $[\mathbf{z}^{(1)}, \mathbf{z}^{(2)}, \cdots, \mathbf{z}^{(N)}]$ and query patches sequence $[\mathbf{z}^{(q_1)}, \mathbf{z}^{(q_2)},\cdots, \mathbf{z}^{(q_Q)}]$. Let $\mathbf{z}' = [\mathbf{z}^{[CLS]}, \mathbf{z}^{(1)}, \mathbf{z}^{(2)}, \cdots, \mathbf{z}^{(N)}]$ and $\mathbf{z}^{q} = [\mathbf{z}^{(q_1)}, \mathbf{z}^{(q_2)},\cdots, \mathbf{z}^{(q_Q)}]$. For $\mathbf{z}^{[CLS]}$ and raw patches, the propagation rule in the attention block is in line with vision transformer, i.e.,
\begin{equation}
\label{eqn:attn}
    \operatorname{Attn} (\mathbf{Q}_{z'}, \mathbf{K}_{z'}, \mathbf{V}_{z'}) = \operatorname{Softmax}\left(\frac{\mathbf{Q}_{z'}\mathbf{K}_{z'}^{\top}}{\sqrt{d_k}}\right) \mathbf{V}_{z'},
\end{equation}
where $\mathbf{Q}_{\mathbf{z}'} = \mathbf{z}'\mathbf{W}_{Q}, \mathbf{K}= \mathbf{z}'\mathbf{W}_{K}, \mathbf{V}= \mathbf{z}'\mathbf{W}_{V}$, with learnable weights $\mathbf{W}_{Q},  \mathbf{W}_{K}, \mathbf{W}_{V}$. For query patches $\mathbf{z}^{(q_i)}$, the attention propagation rule is similar to $[CLS]$ token, i.e.,
\begin{equation}
\label{eqn:queryattn}
    \operatorname{Attn} (\mathbf{Q}_{z^{(q_i)}}, \mathbf{K}_{z'}, \mathbf{V}_{z'}) = \operatorname{Softmax}\left(\frac{\mathbf{Q}_{z}^{(q_i)} \mathbf{D} \mathbf{K}_{z'}^{\top}}{\sqrt{d_k}}\right) \mathbf{V}_{z'},
\end{equation}
where $\mathbf{Q}_{z}^{(q_i)} = \mathbf{z}^{(q_i)}\mathbf{W}_{Q}$ and $\mathbf{D}$ is the filter module as:
\begin{equation}
\begin{split}
 \mathbf{D}_{ij} = \left \{
\begin{array}{lr}
 1, & \vert i-j \vert \leq 1\\
 0, & \vert i-j \vert > 1
\end{array}
\right.
\end{split}
\end{equation} 

\begin{table*}[tb!]
\centering
      \caption{Average accuracy (\%) of each object on Pascal VOC.}
\resizebox{\textwidth}{!}{
\begin{tabular}{r || c c c c c c c c c c c c c c c c c c c c |c}
     	\hline
         model & aero & bike & bird &  boat & bottle & bus & car & cat & chair & cow & table & dog & horse & motor & person & plant & sheep & sofa & train & tv & Avg\\
         \hline
         GMN~\cite{ZanfirCVPR18} & 31.9  & 47.2  & 51.9  & 40.8  & 68.7  & 72.2  & 53.6  & 52.8  & 34.6  & 48.6  & 72.3  & 47.7  & 54.8  & 51.0  & 38.6  & 75.1  & 49.5  & 45.0  & 83.0  & 86.3  & 55.3  \\
         PCA~\cite{WangICCV19} & 49.8  & 61.9  & 65.3  & 57.2  & 78.8  & 75.6  & 64.7  & 69.7  & 41.6  & 63.4  & 50.7  & 67.1  & 66.7  & 61.6  & 44.5  & 81.2  & 67.8  & 59.2  & 78.5  & 90.4  & 64.8  \\
        NGM~\cite{WangPAMI22} & 50.1  & 63.5  & 57.9  & 53.4  & 79.8  & 77.1  & 73.6  & 68.2  & 41.1  & 66.4  & 40.8  & 60.3  & 61.9  & 63.5  & 45.6  & 77.1  & 69.3  & 65.5  & 79.2  & 88.2  & 64.1  \\
        IPCA~\cite{WangICCV19} & 53.8  & 66.2  & 67.1  & 61.2  & 80.4  & 75.3  & 72.6  & 72.5  & 44.6  & 65.2  & 54.3  & 67.2  & 67.9  & 64.2  & 47.9  & 84.4  & 70.8  & 64.0  & 83.8  & 90.8  & 67.7  \\
        CIE~\cite{YuICLR20} & 52.5  & 68.6  & 70.2  & 57.1  & 82.1  & 77.0  & 70.7  & 73.1  & 43.8  & 69.9  & 62.4  & 70.2  & 70.3  & 66.4  & 47.6  & 85.3  & 71.7  & 64.0  & 83.8  & 91.7  & 68.9  \\
        \hline\hline
        BBGM~\cite{rolinek2020deep}	&61.9	&71.1&	79.7	&79	&87.4	&94	&89.5&	80.2	&56.8	&79.1	&64.6	&78.9	&76.2	&75.1	&65.2&	98.2&	77.3	&77&	94.9	&93.9	&79.0\\
        NGMv2~\cite{WangPAMI22}&	61.8&	71.2	&77.6	&78.8	&87.3&	93.6&	87.7&	79.8	&55.4&	77.8	&89.5&	78.8	&80.1&	79.2&	62.6	&97.7&	77.7&	75.7&	96.7&	93.2&	80.1\\
        ASAR~\cite{ren2022appearance} & 62.9 & 74.3 & 79.5 & 80.1 & 89.2 & 94.0 & 88.9 & 78.9 & 58.8 & 79.8 & 88.2 & 78.9 & 79.5 & 77.9 & 64.9 & 98.2 & 77.5 & 77.1 & 98.6 & 93.7 & 81.1 \\
        COMMON~\cite{lin2022graph} & 65.6 & 75.2 & 80.8 & 79.5 & 89.3 & 92.3 & 90.1 & 81.8 & 61.6 & 80.7 & 95.0 & 82.0 & 81.6 & 79.5 & 66.6 & 98.9 & 78.9 & 80.9 & 99.3 & 93.8 & 82.7 \\
        \rowcolor{gray!40} GMTR (Ours) & 69.0 &	74.2&	84.1&	75.9& 87.7 & 94.2&	90.9&	87.8	&62.7	&83.5	&93.9	&84.0&	78.7&	79.6	&69.2&	99.3&	82.5&	83.0&	99.1	&93.3	& \textbf{83.6}\\
        \hline
\end{tabular}}
\vspace{-10pt}
\label{tab:voc}
\end{table*}

\begin{table*}[tb!]
\centering
      \caption{Average accuracy (\%) of each object on SPair-71k.}
\resizebox{\textwidth}{!}{
\begin{tabular}{r || c c c c c c c c c c c c c c c c c c |c}
     	\hline
          model & aero & bike & bird &  boat & bottle & bus & car & cat & chair & cow & dog & horse & motor & person & plant & sheep& train & tv & Avg\\
         \hline
GMN~\cite{ZanfirCVPR18} & 59.9 & 51.0 & 74.3 & 46.7 & 63.3 & 75.5 & 69.5 & 64.6 & 57.5 & 73.0 & 58.7 & 59.1 & 63.2 & 51.2 & 86.9 & 57.9 & 70.0 & 92.4 & 65.3\\
PCA~\cite{WangICCV19} & 64.7 & 45.7 & 78.1 & 51.3 & 63.8 & 72.7 & 61.2 & 62.8 & 62.6 & 68.2 & 59.1 & 61.2 & 64.9 & 57.7 & 87.4 & 60.4 & 72.5 & 92.8 & 66.0\\
NGM~\cite{WangPAMI22} & 66.4 & 52.6 & 77.0 & 49.6 & 67.7 & 78.8 & 67.6 & 68.3 & 59.2 & 73.6 & 63.9 & 60.7 & 70.7 & 60.9 & 87.5 & 63.9 & 79.8 & 91.5 & 68.9\\
IPCA~\cite{WangICCV19} & 69.0 & 52.9 & 80.4 & 54.3 & 66.5 & 80.0 & 68.5 & 71.4 & 61.4 & 74.8 & 66.3 & 65.1 & 69.6 & 63.9 & 91.1 & 65.4 & 82.9 & 97.5 & 71.2\\
CIE~\cite{YuICLR20} & 71.5 & 57.1 & 81.7 & 56.7 & 67.9 & 82.5 & 73.4 & 74.5 & 62.6 & 78.0 & 68.7 & 66.3 & 73.7 & 66.0 & 92.5 & 67.2 & 82.3 & 97.5 & 73.3\\
\hline\hline
BBGM~\cite{rolinek2020deep} & 72.5 & 64.6 & 87.8 & 75.8 & 69.3 & 94.0 & 88.6 & 79.9 & 74.6 & 83.2 & 78.8 & 77.1 & 76.5 & 76.3 & 98.2 & 85.5 & 96.8 & 99.3 & 82.1\\
NGMv2~\cite{WangPAMI22} & 68.8 & 63.3 & 86.8 & 70.1 & 69.7 & 94.7 & 87.4 & 77.4 & 72.1 & 80.7 & 74.3 & 72.5 & 79.5 & 73.4 & 98.9 & 81.2 & 94.3 & 98.7 & 80.2\\
ASAR~\cite{ren2022appearance} & 72.4 & 61.8 & 91.8 & 79.1 & 71.2 & 97.4 & 90.4 & 78.3 & 74.2 & 83.1 & 77.3 & 77.0 & 83.1 & 76.4 & 99.5 & 85.2 & 97.8 & 99.5 & 83.1 \\
COMMON~\cite{lin2022graph} & 77.3 & 68.2 & 92.0 & 79.5 & 70.4 & 97.5 & 91.6 & 82.5 & 72.2 & 88.0 & 80.0 & 74.1 & 83.4 & 82.8 & 99.9 & 84.4 & 98.2 & 99.8 & \textbf{84.5} \\
\rowcolor{gray!40}GMTR (Ours) & 75.6 & 67.2 & 92.4 & 76.9 & 69.4 & 94.8 & 89.4 & 77.5 & 72.1 & 86.3 & 77.5 & 72.2 & 86.4 & 79.5 & 99.6 & 84.4 & 96.6 & 99.7 & 83.2\\

\hline
\end{tabular}}
\vspace{-10pt}
\label{tab:71k}
\end{table*}

For illustration, we visualize the construction of filter matrix $\mathbf{D}$ for one specific query point $\mathcal{P}$ in Fig.~\ref{fig:illufilter}.

By Eq.~\ref{eqn:attn} and Eq.~\ref{eqn:queryattn}, we complete the attention block in the modified transformer. By Eq.~\ref{eqn:queryattn}, we derive the keypoint embeddings $[\mathbf{Z}^{(q1)}, \mathbf{Z}^{(q2)}, \cdots, \mathbf{Z}^{(qQ)}]$. To fully use the spatial information, we also use bilinear algorithm in the raw patch embeddings $\mathbf{Z}^{b} = [\mathbf{Z}^{(b1)}, \mathbf{Z}^{(b2)}, \cdots, \mathbf{Z}^{(bQ)}]$ of the final output layer, as also commonly used in previous GM methods~\cite{WangICCV19}. Then, the final keypoint embeddings are:
$
    \mathbf{Z} = \operatorname{Concat} (\mathbf{Z}^{q}, \mathbf{Z}^{b})
$.

\vspace{-7pt}
\subsection{Backend: GM Solver Transformer}\vspace{-7pt}
\label{sec:backend}
\subsubsection{Association matrix learning}
\vspace{-7pt}
In QAP, the association matrix contains the first order (node-to-node similarity) and second order (edge-to-edge similarity) information of two graphs. We follow the method of~\cite{WangPAMI22} to construct the association matrix with the keypoint features extracted from the frontend module (e.g. QueryTrans). We derive the node embeddings from the association matrix (i.e. association matrix learning) with the backend modules. In contrast with the previous works~\cite{WangICCV19, WangPAMI22, rolinek2020deep}, we apply the graph transformer TransformerConv~\cite{MP_SSL} for association matrix learning, since TransformerConv has shown strengths on both graph-level and node-level prediction tasks over graph convolution networks~\cite{KipfICLR17} and other transformer graph networks~\cite{velivckovic2017graph, brody2021attentive}. Small parameters and simple yet effective pipeline furthermore improve the transferability to combinatorial optimization problem. Specifically, we use $\mathbf{K}_p \in \mathbb{R}^{n_1\times n_2\times d}$ and $\mathbf{K} \in \mathbb{R}^{n_1n_2\times n_1n_2\times d}$ to denote the node similarity matrix and the association matrix, respectively. we view $\operatorname{vec}(\mathbf{K}_p)$ as the node feature, and $\mathbf{K}$ as the edge attribute in TransformerConv. The attention in each layer is calculated as:
\begin{equation}
\begin{split}
     \mathbf{attn}^l_{ij}=\operatorname{Softmax}_{j\in \mathcal{N}(i)}\left(\frac{\mathbf{q}^l_i({\mathbf{k}^l_j}^\top+\mathbf{e}_{ij})}{\sqrt{d}}\right)
\end{split}
\end{equation}
where $\mathcal{N}(i)$ denotes the neighbor of node $i$, namely $\mathbf{W}_{ij}>0$ if $j\in \mathcal{N}(i)$. $\mathbf{q}^l=\mathbf{z}^l\mathbf{W}_Q^l,\mathbf{k}^l=\mathbf{z}^l\mathbf{W}_K^l, \mathbf{e}^l=\mathbf{K}\mathbf{W}_e^l$, $\mathbf{z}^l$ is the node feature of the $l$-th TransformerConv layer.

We make message aggregation for each node as follows:
\begin{equation}
    \mathbf{z}^{l+1}_i=\operatorname{ReLU}(\operatorname{LayerNorm}(\mathbf{attn}^l_{ij} \cdot (\mathbf{v}_j^l+\mathbf{e}_{ij}^l)))
\end{equation}
where, $\mathbf{v}^l = \mathbf{z}^l\mathbf{W}_V^l$.
Specifically, for the last layer, we adopt multi-head attention mechanism, where the activation function and normalization layer are abandoned. 
\begin{equation}
    \mathbf{z}^{k}_i=\frac{1}{H}\sum_{i=1}^C \operatorname{attn}^l_{c, ij}(\mathbf{v}_{c, j}^l+\mathbf{e}_{c, ij}^l)
    \label{eqn:TransformerConvfinal}
\end{equation}
\subsubsection{Keypoint matching}
\vspace{-7pt}
We obtain the node embedding $\mathbf{z}^{final}\in \mathbb{R}^{n_1n_2\times d}$ by Eq.~\ref{eqn:TransformerConvfinal}. We project the embedding  with an FC layer: $\mathbf{m}=\mathbf{z}^{final}\times \mathbf{W}_{proj}$, where $\mathbf{W}_{proj}\in \mathbb{R}^{d\times 1}$ is a learnable parameter. Inspired by~\cite{ZanfirCVPR18, WangICCV19}, we adopt a Sinkhorn network~\cite{SinkhornAMS64} $f: \mathbb{R}^{n_1n_2\times 1}\to [0, 1]^{n_1\times n_2}$ to calculate the matching. 

As each entry in matching matrix can be regarded as a binary classification where $1$ denotes matched and $0$ denotes unmatched, we adopt binary cross-entropy as our loss function. Given matching ground truth $\mathbf{X}^{gt}$, the loss is:
\begin{equation}
\small{
    \mathcal{L}=-\frac{\sum_{i=1}^{n_1}\sum_{j=1}^{n_2}\left(\mathbf{X}^{gt}_{ij}\log(\mathbf{m}_{ij}) + (1-\mathbf{X}^{gt}_{ij})\log(1-\mathbf{m}_{ij})\right)}{n_1 n_2}
} \notag
\end{equation}

Please note that our backend module can solve the QAP independently without the front-end module since the only input is the QAP affinity matrix, which is, to our best knowledge, the first successful application for transformer in Lawler's Quadratic Assignment Problem (QAP).

\begin{table}[tb!]
\centering
  \caption{Average accuracy (\%) with combination of backbones under two baseline frameworks: NGMv2 and BBGM.}
\resizebox{0.75\columnwidth}{!}{
\begin{tabular}{r c c c c}
    \toprule
    \multirow{2}{*}{Backbone} & \multicolumn{2}{c}{Pascal VOC} & \multicolumn{2}{c}{SPair-71k} \\
    \cmidrule(lr){2-3}
    \cmidrule(lr){4-5}
    & NGMv2 & BBGM & NGMv2 & BBGM \\
    \midrule
    VGG16~\cite{simonyanICLR14vgg} & 80.1 & 79.0 & 80.6 & 82.1\\
    ResNet34~\cite{he2016deep} & 80.2 & 80.7 &80.4 & 81.5\\
    ViT~\cite{vit} & 82.7 & 83.6 & 80.8 & 83.0\\
    CeiT~\cite{yuan2021incorporating} & 82.6 & 83.1 & 81.6 & 81.4\\
    XCiT~\cite{ali2021xcit} & 82.7 & 82.8 & 81.7 & 82.3 \\
    \rowcolor{gray!40}QueryTrans (Ours) & \textbf{83.3} & \textbf{84.5} & \textbf{82.5} & \textbf{83.9}\\
    \bottomrule
\end{tabular}
}
\vspace{-15pt}
\label{tab:backbone}
\end{table}
\vspace{-7pt}
\section{Experiments}
\vspace{-7pt}
\subsection{Overall Performance on Pascal VOC}\vspace{-7pt}
We conduct experiments on Pascal VOC, which consists of 20 instance classes, with each image containing 6 to 23 key points. To compare our model with peer approaches, we follow the same settings as previous works~\cite{WangICCV19, WangPAMI22, rolinek2020deep}, which involve filtering the outlier points to guarantee only the points present in both source image and target image are preserved. Images are then re-scaled according to the frontend. Results are given in Table~\ref{tab:voc}. GMTR achieves the SOTA performance.
\vspace{-7pt}
\subsection{Overall Performance on SPair-71k}\vspace{-7pt}
SPair-71k is a large dataset consisting of 70,958 image pairs with diverse variations in viewpoint and scale. Compared to Pascal VOC dataset, it is large in number and contains more annotations. We follow the same setting as in the previous section, select intersection points, and each image is cropped to its bounding box and scaled to $256\times 256$ or $224\times 224$. Table~\ref{tab:71k} indicates that GMTR shows strengths on SPair-71k as well, outperforming most of the peer works.

\vspace{-7pt}
\subsection{Ablation Study}\vspace{-7pt}
\subsubsection{Study on Backbones}\vspace{-7pt}
We test CNN-based and transformer-based backbones under the NGMv2 and BBGM frameworks on Pascal VOC  and SPair-71k. For CNN-based, we test VGG16~\cite{simonyanICLR14vgg} and ResNet34~\cite{he2016deep}, which are two classical models. For transformer-based model, we conduct experiments on ViT~\cite{vit}, CeiT~\cite{yuan2021incorporating} and XCiT~\cite{ali2021xcit}, which show competitive performance on ImageNet1k. Finally, we test our QueryTrans, which adopts both self-attention and cross-attention mechanism to extract information. All other parameters irrelevant to the backbone and image rescale values ($256\times 256$ or $224\times 224$) are kept constant across all models. The results are shown in Table~\ref{tab:backbone}.

\textbf{Results on Pascal VOC dataset}: 
For both NGMv2 and BBGM frameworks, we set learning rate $2\times 10^{-5}$ for backbone training and $2\times 10^{-3}$ for other modules. The backend modules are implemented using a 3-layer GCN~\cite{KipfICLR17}, whose feature dimension is 16. The results show that our proposed QueryTrans performs the best under both frameworks.

\textbf{Results on SPair-71k:}
We test QueryTrans and previous mentioned CNN-based and transformer-based backbones with NGMv2 and BBGM on SPair-71k. The learning rate is set as $1\times 10^{-5}$ for backbones and $1\times 10^{-3}$ for other modules. The results indicate that QueryTrans outperforms all the backbones under both frameworks on SPair-71k dataset.
\vspace{-7pt}
\subsubsection{Study on TransformerConv Modules}\vspace{-7pt}
We conduct experiments to show the strengths of TransformerConv~\cite{MP_SSL} over GNNs (e.g. GCN~\cite{KipfICLR17}). The number of GCN layers is set as 3, with feature dimension of 17 (16 + Sinkhorn~\cite{SinkhornAMS64}). The number of TransformerConv layers is 3, with feature dimension of 16. We adopt multi-head self attention where the number of heads for each layer is 1, 1, and 2 respectively. To verify the necessity of backend module, we also test the performance without backend module. The node information is directly passed to the final layer.

\begin{table}[!tb]
      \caption{Ablation study of the variants of backend modules.}
\resizebox{\columnwidth}{!}{
\begin{tabular}{c|c|c}
\hline
Exp. & Setup (on Pascal VOC dataset) & accuracy \\
\hline
A & baseline(QueryTrans + TransformerConv + positional encoding) & 83.63 \\
B & A - TransformerConv + GCN & 83.28 \\
C & A - TransformerConv & 73.09 \\
D & A - positional encoding & 76.12\\
E & B - positional encoding & 75.08 \\ 
\hline
\end{tabular}}
\vspace{-15pt}
\label{tab:backend}
\end{table}

In Table~\ref{tab:backend}, Exp.~A \& B show that the attention mechanism combined with the association graph is more effective in updating node features and capturing the relationships between nodes and their neighbors. Exp.~C highlights the necessity of the backend module. Exp.~D \& E show that positional encoding is crucial for both GCN and TransformerConv.
\vspace{-7pt}
\subsubsection{Study on Attention with Filter}\vspace{-7pt}
\label{sec:expop_att_filter}

One key design of our QueryTrans is the filter, which masks the raw patches that contain no overlapping area with the query patch. We conduct expoperiments to demonstrate the necessity of the filter. We also expoplore the performance of different node embedding designs. Specifically, \textbf{bilinear} means the model adopts the bilinear algorithm over raw patches from the last two layers of QueryTrans, and concatenates the two vectors as node feature. \textbf{Cross-attention} means the model concatenates the query patches from the last two layers of QueryTrans as the output. \textbf{cross-attention+bilinear} denotes the model adopts the blinear algorithm over raw patches from the last layer of QueryTrans, and concatenate them with query patches from the last layer of QueryTrans as the output. The results are given in Table~\ref{tab:filter}, where \textbf{cross-attention+bilinear+filter} performs the best, indicating that local query information combined with spatial information provides more accurate features.

\begin{table}[tb!]
  \renewcommand{\arraystretch}{1.}
      \caption{Ablation study of image information extract method based on vision transformer on Pascal VOC dataset.}
    \vspace{5pt}
\resizebox{0.9\columnwidth}{!}{
\begin{tabular}{r|c|c}
\toprule
Method & Accuracy & Relative acc \\
\hline
bilinear (ViT) & 82.68 & 0 \\
cross-attention & 77.98 & -4.70  \\
cross-attention + filter & 81.88 & -0.80\\
bilinear + corss-attention& 79.26 & -3.42  \\
bilinear + cross-attention  + filter & 83.63 & +0.95 \\
\bottomrule
\end{tabular}}
\label{tab:filter}
\vspace{-10pt}
\end{table}

\vspace{-7pt}
\section{Conclusion}
\vspace{-7pt}
We have proposed GMTR, a transformer-based approach for visual graph matching, with two key components, i.e., QueryTrans and a backend transformer, where QueryTrans learns the visual keypoints information from keypoint crops and cross attention mechanism, and the backend transformer further improves the matching accuracy. Moreover, QueryTrans can be integrated into popular GM frameworks (NGMv2, BBGM). Extensive experimental results on popular GM benchmarks show that GMTR achieves competitive performance.

\bibliographystyle{IEEEbib}
\bibliography{eigbib}

\end{document}